\documentclass[11pt]{article} 
\usepackage{amsfonts,amsmath,amssymb,amsthm,url,xspace}
\usepackage{graphicx,subfigure}
\usepackage{fullpage}
\usepackage{natbib}
\usepackage{authblk}
\usepackage[pointedenum]{paralist}
\usepackage[ruled,noline]{algorithm2e}
\usepackage{epic,eepic,eepicemu}
\usepackage{epsfig}
\usepackage{fancyhdr}
\usepackage{graphics}
\usepackage{psfrag,latexsym}
\usepackage{multirow}
\usepackage{tabularx}
\usepackage{epsfig}
\theoremstyle{break}

\usepackage{enumitem}

\newcommand{\argmax}{\operatornamewithlimits{argmax}}
\newcommand{\argmin}{\operatornamewithlimits{argmin}}

\def\ml1m{{\sf movielens1m}\xspace}

\def\movielens10m{{\sf movielens10m}\xspace}

\def\HepPh{{\sf ca-HepPh}\xspace}
\def\GrQc{{\sf ca-GrQc}\xspace}
\def\LiveJournal{{\sf LiveJournal}\xspace}
\def\MySpace{{\sf MySpace}\xspace}

\newcommand{\Proj}{{\mathcal{P}}}
\def\D{\mathcal{D}}
\def\W{\mathcal{W}}
\def\S{\mathcal{S}}

\def\R{\mathbb{R}}
\def\X{\mathcal{X}}
\def\S{\mathcal{S}}

\def\bh{{\boldsymbol h}}

\def\bx{{\boldsymbol x}}

\def\bv{{\boldsymbol v}}
\def\bu{{\boldsymbol u}}

\def\be{{\boldsymbol e}}

\newtheorem{lemma}{Lemma}

\newtheorem{theorem}{Theorem}

\begin{document}
\title{PU Learning for Matrix Completion}
\author[1]{Cho-Jui Hsieh}
\author[1]{Nagarajan Natarajan} 
\author[1]{Inderjit S. Dhillon}
\affil[1]{Department of Computer Science, University of Texas, Austin}

\renewcommand\Authands{ and }
\date{}

\maketitle

\begin{abstract}
In this paper, we consider the matrix completion problem when the observations are one-bit measurements of some underlying matrix $M$, and in particular the observed samples consist \emph{only} of ones and no zeros. This problem is motivated by modern applications such as recommender systems and social networks where only ``likes'' or ``friendships'' are observed. The problem of learning from only positive and unlabeled examples, called PU (positive-unlabeled) learning, has been studied in the context of binary classification. We consider the PU matrix completion problem, where an underlying real-valued matrix $M$ is first quantized to generate one-bit observations and then a subset of positive entries is revealed. Under the assumption that $M$ has bounded nuclear norm, we provide recovery guarantees for two different observation models: 1) $M$ parameterizes a distribution that generates a binary matrix, 2) $M$ is thresholded to obtain a binary matrix. For the first case, we propose a ``shifted matrix completion'' method that recovers $M$ using only a subset of indices corresponding to ones, while for the second case, we propose a ``biased matrix completion'' method that recovers the (thresholded) binary matrix. Both methods yield strong error bounds --- if $M \in \mathbb{R}^{n\times n}$, the Frobenius error is bounded as $O\big(\frac{1}{(1-\rho)n}\big)$, where $1-\rho$ denotes the fraction of ones observed. This implies a sample complexity of $O(n \log n)$ ones to achieve a small error, when $M$ is dense and $n$ is large. We extend our methods and guarantees to the recently proposed inductive matrix completion problem, where rows and columns of $M$ have associated features. We provide efficient and scalable optimization procedures for both the methods and demonstrate the effectiveness of the proposed methods for link prediction (on real-world networks consisting of over 2 million nodes and 90 million links) and semi-supervised clustering tasks.
\end{abstract}

\section{Introduction}
The problem of recovering a matrix from a given subset of its entries arises in many practical problems of interest. The famous Netflix problem of predicting user-movie ratings is one example that motivates the traditional matrix completion problem, where we would want to recover the underlying (ratings) matrix given partial observations. Strong theoretical guarantees have been developed in the recent past for the low-rank matrix completion problem~\citep{EJC09a,Candes:09,candes2010power}. An important variant of the matrix completion problem is to recover an underlying matrix from one-bit quantization of its entries. Modern applications of the matrix completion problem reveal a conspicuous gap between existing matrix completion theory and practice. For example, consider the problem of link prediction in social networks. Here, the goal is to recover the underlying friendship network from a given snapshot of the social graph consisting of observed friendships. We can pose the problem as recovering the adjacency matrix of the network $A$ such that $A_{ij} = 1$ if users $i$ and $j$ are related and $A_{ij} = 0$ otherwise. In practice, we only observe positive relationships between users corresponding to 1's in $A$. Thus, there is not only one-bit quantization in the observations, but also a one-sided nature to the sampling process here --- no ``negative'' entries are sampled. In the context of classification, methods for learning in the presence of positive and unlabeled examples only, called positive-unlabeled (PU in short) learning, have been studied in the past~\citep{elkan2008learning, liu2003building}. For matrix completion, can one guarantee recovery when only a subset of positive entries is observed? In this paper, we formulate the PU matrix completion problem and answer the question in the affirmative under different settings.

Minimizing squared loss on the observed entries corresponding to 1's, subject to the low-rank constraints, yields a degenerate solution --- the rank-1 matrix with all its entries equal to 1 achieves zero loss. In practice, a popular heuristic used is to try and complete the matrix by treating some or all of the missing observations as true 0's, which seems to be a good strategy when the underlying matrix has a small number of positive examples, i.e., small number of 1's. This motivates viewing the problem of learning from only positive samples as a certain noisy matrix completion problem. Existing theory for noise-tolerant matrix completion \citep{EJC09a,onebit} does not sufficiently address recoverability under PU learning (see Section \ref{sec:settings}).  

In our work, we assume that the true matrix $M \in \mathbb{R}^{m \times n}$ has a bounded nuclear norm $\|M\|_{*}$. The PU learning model for matrix completion is specified by a certain one-bit quantization process that generates a binary matrix $Y$ from $M$ and a one-sided sampling process that reveals a subset of positive entries of $Y$. In particular, we consider two recovery settings for PU matrix completion: The first setting is non-deterministic --- $M$ parameterizes a probability distribution which is used to generate the entries of $Y$. We show that it is possible to recover $M$ using only a subset of positive entries of $Y$. The idea is to minimize an unbiased estimator of the squared loss between the estimated and the observed ``noisy'' entries, motivated by the approach in \cite{naga_nips}. We recast the objective as a ``shifted matrix completion'' problem that facilitates in obtaining a scalable optimization algorithm. The second setting is deterministic --- $Y$ is obtained by thresholding the entries of $M$ (modeling how the users vote), and then a subset of positive entries of $Y$ is revealed. While recovery of $M$ is not possible (see Section \ref{sec:settings}), we show that we can recover $Y$ with low error. To this end, we propose a scalable biased matrix completion method where the observed and the unobserved entries of $Y$ are penalized differently. Recently, an inductive approach to matrix completion was proposed~\citep{dhillon_inductive} where the matrix entries are modeled as a bilinear function of real-valued features associated with the rows and the columns. We extend our methods under the two aforementioned settings to the inductive matrix completion problem and establish similar recovery guarantees. Our contributions are summarized below:
\begin{compactenum}
\item To the best of our knowledge, this is the first paper to formulate and study PU learning for matrix completion, necessitated by the applications of matrix completion. Furthermore, we extend our results to the recently proposed inductive matrix completion problem.
\item We provide strong guarantees for recovery; for example, in the non-deterministic setting, the error in recovering an $n \times n$ matrix is $O(\frac{1}{(1-\rho)n})$ for our method compared to $O(\frac{1}{(1-\rho)\sqrt{n}})$ implied by the method in \cite{onebit}, where $(1-\rho)$ is the fraction of observed 1's. 
\item Our results provide a theoretical insight for the heuristic approach used in practice, namely, biased matrix completion.
\item We give efficient, scalable optimization algorithms for our methods; experiments on simulated and real-world data (social networks consisting of over 2 million users and 90 million links) demonstrate the superiority of the proposed methods for the link prediction problem.
\end{compactenum}

\paragraph{Outline of the paper.} We begin by establishing some hardness results and describing our PU learning settings in Section \ref{sec:settings}. In Section \ref{sec:MC}, we propose methods and give recovery guarantees for the matrix completion problem under the different settings. We extend the results to PU learning for inductive matrix completion problem in Section \ref{sec:IMC}. We describe efficient optimization procedures for the proposed methods in Section \ref{sec:optimization}. Experimental results on synthetic and real-world data are presented in Section \ref{sec:experiments}. 

\paragraph{Related Work. }
In the last few years, there has been a tremendous amount of work on the theory of matrix completion since the remarkable result concerning recovery of low-rank matrices by \cite{Candes:09}. Strong results on recovery from noisy observations have also been established~\citep{EJC09a,keshavan2010matrix}. Recently, \cite{onebit} studied the problem of recovering matrices from 1-bit observations, motivated by the nature of observations in domains such as recommender systems where matrix completion is heavily applied. Our work draws motivation from recommender systems as well, but differs from \cite{onebit} in that we seek to understand the case when only 1's in the matrix are observed. One of the algorithms we propose for PU matrix completion is based on using different costs in the objective for observed and unobserved entries. The approach has been used before, albeit heuristically, in the context of matrix completion in recommender system applications~\citep{VS10a}. 
Compressed sensing is a field that is closely related to matrix completion. Here the goal is to recover an $s$-sparse vector in $\mathbb{R}^{d}$ using a limited number of linear measurements. Recently, compressed sensing theory has been extended to the case of single-bit quantization~\citep{boufounos20081}. Here, the goal is to recover an $s$-sparse signal when the observations consist of only the signs of the measurements, and remarkable recovery guarantees have been proved for the single-bit quantization case~\citep{boufounos20081}.

%

\section{Problem Settings}
\label{sec:settings}
We assume that the underlying matrix $M\in \R^{m\times n}$ has a bounded nuclear norm, i.e., $\|M\|_*\leq t$, where $t$ is a constant independent of $m$ and $n$. If $M_{ij}\in \{0,1\}$ for all $(i,j)$, stating the PU matrix completion problem is straight-forward: we only observe a subset $\Omega_1$ randomly sampled from $\{(i,j)\mid M_{ij}=1\}$ and the goal is to recover $M$ based on this ``one-sided'' sampling. We call this the ``basic setting''.  However, in real world applications 
it is unlikely that the underlying matrix is binary. In the following, we consider two general 
settings, which include the basic setting as a special case.

\subsection{Non-deterministic setting}
\label{sec:setting-random}
In the non-deterministic setting, we assume $M_{ij}$ has bounded values and without loss of generality we can assume $M_{ij}\in [0, 1]$ for all $(i,j)$ by normalizing it. We then consider each entry as a probability distribution which generates a {\bf clean} 0-1 observation $Y\in \R^{m\times n}$: 
 \begin{equation*}
   P(Y_{ij}=1) = M_{ij}, \ \ P(Y_{ij}=0) = 1-M_{ij}, 
 \end{equation*}
 In the classical matrix completion setting, we will observe partial entries sampled randomly from $Y$; In our PU learning model, we assume only a subset of positive entries of $Y$ is observed.  More precisely, we observe a subset $\Omega_1$ from $Y$ where $\Omega_1$ is sampled uniformly from 
 $\{(i,j) \mid Y_{ij}=1\}$. We assume $|\Omega_1|=\bar{s}$ and denote the number of $1$'s in $Y$ by $s$. 
 With only $\Omega_1$ given, the goal of PU matrix completion is to recover the underlying
 matrix $M$.  Equivalently, letting $A\in \{0,1\}^{m\times n}$ to denote the observations, 
 where $A_{\Omega_1}=1$ and $A_{ij}=0$ for all $(i,j)\notin \Omega_1$, the non-deterministic setting can be specified as observing $A$ by the process:
\begin{equation}
   P(A_{ij}=1) = M_{ij}(1-\rho), \ \ 
   P(A_{ij}=0) = 1-M_{ij}(1-\rho), 
   \label{eq:random_A}
 \end{equation}
 where $\rho=1 - \bar{s}/s$ is the noise rate of flipping a 1 to 0 (or equivalently, $1-\rho$ is the sampling ratio to obtain $\Omega_{1}$ from $Y$).

\paragraph{Hardness of recovering $M$:}
The 1-bit matrix completion approach of \cite{onebit} can be applied to this setting ---  Given a matrix $M$, a subset $\Omega$ is sampled uniformly at random from $M$, and the observed values are ``quantized'' 
 by a known probability distribution. We can transform our problem to the 1-bit matrix completion problem by assuming all the unobserved entries are zeros. For convenience we assume $M\in \R^{n\times n}$. 
%
We will show that the one-bit matrix completion approach in \citep{onebit} is not satisfactory for PU matrix completion in the non-deterministic setting. In \citep{onebit}, the underlying matrix $M$ is assumed to satisfy $\|M\|_{*} \leq t$ and $\|M\|_\infty\leq \alpha$; we are given a subset (chosen uniformly random) $\Omega$ with
 $|\Omega|=m$  and we observe the following quantity on $\Omega$:
 \begin{equation}
   Y_{i,j} = \begin{cases}
     1 &\text{ with probability } f(M_{ij}), \\
     -1 & \text{ with probability } 1-f(M_{ij}).  
   \end{cases}
   \label{eq:onebit-generate}
 \end{equation}
 By setting $f(M_{ij})=(1-\rho)M_{ij}$ and $1\geq M_{ij}\geq 0$, and assuming $\Omega$ contains
 all the $n^2$ entries, it is equivalent to our problem. 

The estimator is obtained by solving the following optimization problem: 
 \begin{equation}
   \hat{M} = \argmax_{X: \|X\|_*\leq t} \sum_{i,j\in \Omega} \left( \sum_{i,j: Y_{ij}=1} \log(f(X_{ij})) 
   + \sum_{i,j: Y_{ij}=0} \log(1-f(X_{ij}))\right).
   \label{eq:onebit_opt}
 \end{equation}
The following result shows that $\hat{M}$ is close to $M$:
\begin{theorem}[\cite{onebit}]
  Assume $\|M\|_*\leq t$, and $Y$ is generated by \eqref{eq:onebit-generate}, 
  then 
  \begin{equation}
    \frac{1}{n^2} \|\hat{M}-M\|_F^2 \leq \sqrt{2} C_\alpha \sqrt{\frac{2nr}{m}}, 
  \end{equation}
  where $m=|\Omega|$, 
  $C_\alpha :=C_2 \alpha L_\alpha \beta_\alpha$ where $C_2$ is a constant and 
  \begin{equation*}
    L_\alpha = \sup_{|x|\leq \alpha} \frac{|f'(x)|}{f(x)(1-f(x))} \text{ and }
    \beta_\alpha = \sup_{|x|\leq \alpha} \frac{f(x)(1-f(x))}{(f'(x))^2}. 
  \end{equation*}
\end{theorem}
By substituting $f(x)=(1-\rho)x$ into the above formulas we can find 
$L_{\alpha} \geq \frac{1}{\alpha(1-(1-\rho)\alpha)}$ and $\beta_\alpha \geq \frac{\alpha(1-(1-\rho)\alpha)}{1-\rho}$, so $L_\alpha \beta_\alpha \geq \frac{1}{1-\rho}$. Therefore the above theorem suggests that 
\begin{equation*}
  \frac{1}{n^2}\|\hat{M}-M\|_F^2 \leq O\bigg(\frac{\sqrt{nr}}{(1-\rho)\sqrt{m}}\bigg). 
\end{equation*}
In our setting, $m=n^2$, so we have
\begin{equation}
  \frac{1}{n^2} \|\hat{M}-M\|_F^2 \leq O\bigg(\frac{\sqrt{r}}{(1-\rho)\sqrt{n}}\bigg). 
  \label{eqn:onebithardness}
\end{equation}
Thus the recovery error is $O\big(\frac{1}{(1-\rho)\sqrt{n}}\big)$, which implies that the sample complexity for recovery using this approach is quite high: For example, observing $O(n \log n)$ 1's, when $M$ is dense, is not sufficient.

The main drawback of using this approach for PU matrix completion is computation --- time complexity of solving \eqref{eq:onebit_opt} is $O(n^2)$ which makes the approach prohibitive for large matrices. Moreover, the average error on each element is $O(1/\sqrt{n})$ (in contrast, our algorithm has $O(1/n)$ average error). To see how this affects sample complexity for recovery, assume $\sum_{i,j}M_{i,j} = O(n^2)$ (number of 1's are of the same order as the number of 0's in the original matrix) and $O(n\log n)$ 1's are observed. Then $(1-\rho) = O(\frac{\log n}{n})$ and the average error according to \eqref{eqn:onebithardness} is $\frac{1}{n^2} \|\hat{M}-M\|_F^2 = O(\frac{\sqrt{rn}}{\log n})$, which diverges as $n\rightarrow \infty$. In contrast, we will show that the average error of our estimator vanishes as $n\rightarrow \infty$. 

%
%

 \subsection{Deterministic setting} 
 \label{sec:deterministic}

 In the deterministic setting,
a {\bf clean} 0-1 matrix $Y$ is observed from $M$ by the thresholding process: 
$Y_{ij} = I(M_{ij}>q)$, where $I(\cdot)$ is the indicator function and $q\in \R$ 
is the threshold. 
Again, in our PU learning model, we assume only a subset of positive entries of $Y$ are observed, i.e. we 
 observe $\Omega_1$ from $Y$ where $\Omega_1$ is sampled uniformly from $\{(i,j) \mid Y_{ij}=1\}$. Equivalently, we will use $A$ to denote the observations, where $A_{ij}=1 $ if $(i,j)\in \Omega_1$, and $A_{ij}=0$ otherwise. 

 It is impossible to recover $M$ even if we observe all the entries of $Y$. A trivial 
 example is that all the matrices $\eta \be\be^T$ will give $Y=\be\be^T$ if $\eta>q$, 
 and we cannot recover $\eta$ from $Y$.   Therefore, in the deterministic setting we can only hope to recover the underlying 0-1 matrix   $Y$ from the given observations. To the best of our knowledge, there is no existing work that gives a reasonable guarantee of recovering $Y$.  For example, if we apply the noisy matrix completion algorithm proposed in \citep{EJC09a}, the estimator $\hat{Y}$ has an error bound $\|\hat{Y}-Y\| \leq \|A-Y\|$, which indicates the error in $\hat{Y}$  is not guaranteed to be better than the trivial estimator $A$.
 
\paragraph{Hardness of applying noisy matrix completion to our deterministic setting:}
An easy way to model PU matrix completion problem in the deterministic setting is to think of it as a traditional matrix completion problem with $n^2$ ``noisy'' observed entries. In \citep{EJC09a}, it is assumed that $A=M+Z$ where $Z$ is noise and $\delta = \|Z\|_F$.  The idea is to solve:
\begin{equation}
  \label{eq:noise_completion}
  \min \|X\|_* \text{ such that } \|\Proj_\Omega (X-A)\|_F \leq \delta, 
\end{equation}
where $\delta$ is total amount of noise. \cite{EJC09a} established the following recovery guarantee: 
\begin{theorem}[\cite{EJC09a}]
  \label{thm:noise_completion}
  Let $M\in \R^{n\times n}$ be a fixed matrix of rank $r$, and assume $M$ is $\mu$-incoherent, i.e.,
  \begin{equation}
    \|\bu_i\|_\infty \leq \sqrt{\mu/n} \text{ and } \|\bv_i\|_\infty \leq \sqrt{\mu/n}, 
  \end{equation}
  where $\bu_i, \bv_i$  are eigenvectors of $M$. 
  Suppose we observe $m$ entries
  of $M$ with locations sampled uniformly at random, and 
  \begin{equation}
    m \geq  C \mu^2 n r \log^6 n, 
  \end{equation}
  where $C$ is a numerical constant, then
  \begin{equation}
    \|\hat{M}-M\| \leq 4\sqrt{\frac{(2+p)n}{p}}\delta + 2\delta, 
  \end{equation}
  where $p=m/n^2$. 
\end{theorem}

If we apply Theorem \ref{thm:noise_completion} to our case, 
$\delta = \|Z\|_F = \frac{\rho}{1-\rho}\bar{s}$, the error of the recovered matrix $\hat{M}$ using \eqref{eq:noise_completion} can be bounded as:
\begin{equation}
  \|\hat{M}-M\|_F \leq \|A-M\|_F,  
  \label{eq:trivial}
\end{equation}
and clearly this bound is not very useful.

\section{Proposed Algorithms for PU Matrix Completion}
\label{sec:MC}

In this section, we introduce two algorithms: shifted matrix completion for non-deterministic PU matrix completion, and biased matrix completion for deterministic PU matrix completion. All proofs are deferred to Appendix \ref{sec:proofs}.

 \subsection{Shifted Matrix Completion for Non-deterministic Setting (ShiftMC)}
 \label{sec:SMC_bound}
We want to find a matrix $X$ such that the loss $\|M-X\|_F^2$ is bounded, using the noisy observation matrix $A$ generated from $M$ by \eqref{eq:random_A}. Observe that conditioned on $Y$, the noise in $A_{ij}$ is \emph{asymmetric}, i.e. $P(A_{ij} = 0 | Y_{ij} = 1) = \rho$ and $P(A_{ij} = 1 | Y_{ij} = 0) = 0$.  Asymmetric label noise has been studied in the context of binary classification, and recently \cite{naga_nips} proposed a method of unbiased estimator to bound the true loss using only noisy observations. In our case, we aim to find 
a matrix minimizing the unbiased estimator defined on each element, which leads to the following optimization problem: 
\begin{equation}
  \label{eq:alg1}
  \min_X \sum_{i,j} \tilde{\ell}(X_{ij}, A_{ij}) \ \text{ such that }
  \|X\|_* \leq t, 0\leq X_{ij} \leq 1 \ \forall (i,j). 
\end{equation}
\begin{equation}
  \text{where} \ \  \tilde{\ell}(X_{ij}, A_{ij}) = \begin{cases}
    \frac{(X_{ij}-1)^2 - \rho X_{ij}^2}{1-\rho}  &\text{ if } A_{ij}=1, \\
    X_{ij}^2 &\text{ if }A_{ij} = 0. 
  \end{cases}
  \label{eq:ltilde_define}
\end{equation}
The bound constraint on $X$ in the above estimator ensures the loss has bounded Lipschitz constant. This optimization problem is equivalent to the traditional trace-norm regularization problem 
\begin{equation}
  \label{eq:alg2}
  \min_X \sum_{i,j} \tilde{\ell}(X_{ij}, A_{ij}) + \lambda \|X\|_*, \text{ such that } 0\leq X_{ij} \leq 1
  \ \forall (i,j),  
\end{equation}
where $\lambda$ has a one-to-one mapping to $t$.  We use $\tilde{\ell}$ instead of the original loss $\ell$ because it is the unbiased estimator of the underlying squared loss $\ell(X_{ij}, M_{ij})= (X_{ij}-M_{ij})^2$,  as formalized below. Thus, we use $\tilde{\ell}$ on the observed $A_{ij}$, we minimize the loss w.r.t. $Y_{ij}$ in expectation.
\begin{lemma}
  For any $X\in \R^{m\times n}$, 
$  \frac{1}{mn} E\big[\sum_{i,j} (X_{ij}-Y_{ij})^2\big] = \frac{1}{mn}E\big[\sum_{i,j}\tilde{\ell}(X_{ij}, A_{ij})\big]$. 
  \label{thm:2}
\end{lemma}
Interestingly, we can rewrite $\tilde{\ell}$ as $\tilde{\ell}(X_{ij}, 1)  
  = \big(X_{ij}-\frac{1}{1-\rho}\big)^2 - \frac{\rho}{(1-\rho)^2}$. Therefore, \eqref{eq:alg2} can be rewritten as the following ``shifted matrix completion'' problem: 
  \begin{equation}
  \label{eq:alg3}
  \hat{X} = \argmin_X \sum_{i,j: A_{ij}=1} \bigg(X_{ij}-\frac{1}{1-\rho}\bigg)^2 + \sum_{i,j: A_{ij}=0} X_{ij}^2 
  + \lambda \|X\|_* \text{ s.t. } 0\leq X_{ij}\leq 1 \ \forall (i,j). 
\end{equation}
We want to show that the average error of the ShiftMC estimator $\hat{X}$ decays as $O(1/n)$. In order to do so, we first need to bound the difference between the expected error and the empirical error. We define the hypothesis space to be $\X:=\{X\mid X\in \R^{m\times n} \text{ and }\|W\|_*\leq t\}$. The expected error can be written as $E_A[R_{\tilde{\ell}}(W)]=E_A[\frac{1}{mn} \sum_{i,j} \tilde{\ell}(W_{ij}, A_{ij})]$, and the empirical error is $\hat{R}_{\tilde{\ell}}(W)=\frac{1}{mn}\sum_{i,j}\tilde{\ell}(W_{ij}, A_{ij})$. We first show that the difference between expected error and empirical error can be upper bounded: 
\begin{theorem}
  Let $\X:=\{X\in \R^{m\times n} \mid \|X\|_* \leq t , 0\leq X\leq 1\}$, then 
  \begin{equation}
    \max_{X\in \X} \Big| E_{A}[R_{\tilde{\ell}}({X})] - \hat{R}_{\tilde{\ell}}(X)\Big|  \leq  tC\frac{\sqrt{n}+\sqrt{m}+\sqrt[4]{s}}{(1-\rho)mn}
    + 3\frac{\sqrt{\log(2/\delta)}}{\sqrt{mn}(1-\rho)}
  \end{equation}
  with probability at least $1-\delta$, 
  where $C$ is a constant, $E[R_{\tilde{\ell}}(X)]:=E[\frac{1}{mn} \sum_{i,j} \tilde{\ell}(X_{ij}, A_{ij})]$ is the expected error, and
$\hat{R}_{\tilde{\ell}}(X)=\frac{1}{mn}\sum_{i,j}\tilde{\ell}(X_{ij}, A_{ij})$ is the empirical
error. 
  \label{thm:mc_empirical_diff}
\end{theorem}
Combining Lemma \ref{thm:2} and Theorem \ref{thm:mc_empirical_diff}, 
we have our first main result:
\begin{theorem} [Main Result 1]
  With probability at least $1-\delta$, 
  \begin{equation*}
    \frac{1}{mn} \sum_{i,j} (M_{ij}-\hat{X}_{ij})^2 \leq 6\frac{\sqrt{\log(2/\delta)}}{\sqrt{mn}(1-\rho)} 
    + 2Ct\frac{\sqrt{n}+\sqrt{m}+\sqrt[4]{s}}{(1-\rho)mn}. 
  \end{equation*}
  \label{thm:mc_random}
\end{theorem}
The average error is of the order of $O(\frac{1}{n(1-\rho)})$ when $M\in \R^{n\times n}$, where $1-\rho$ denotes the ratio of observed 1's. This shows that even when we only observe a very small ratio of 1's in the matrix, we can still estimate $M$ accurately when $n$ is large enough. 

\subsection{Biased Matrix Completion for Deterministic Setting (BiasMC)}
 \label{sec:BMC_bound}
In the deterministic setting, we propose to solve the 
matrix completion problem with label-dependent loss \citep{CS12a}. 
Let $\ell(x, a)=(x-a)^2$ denote the squared loss, for $a \in \{0,1\}$.  The $\alpha$-weighted loss is defined by 
\begin{equation}
  \ell_{\alpha}(x,a) = \alpha 1_{a=1} \ell(x,1) + (1-\alpha) 1_{a=0} \ell(x, 0), 
\end{equation}
where $1_{a=1}, 1_{a=0}$ are indicator functions. We then recover the
groundtruth by solving the following biased matrix completion (biasMC) problem: 
\begin{equation}
  \hat{X} = \argmin_{X: \|X\|_*\leq t} \sum_{i,j} \ell_{\alpha}(X_{ij}, A_{ij})   
  = \argmin_{X: \|X\|_* \leq t} \alpha \sum_{i,j: A_{ij}=1}(X_{ij}-1)^2 + (1-\alpha) \sum_{i,j: A_{ij}=0} X_{ij}^2
  \label{eq:bias-factorization}
\end{equation}
The underlying binary matrix $Y$ is then recovered by the thresholding operator 
$\bar{X}_{ij}=I(\hat{X}_{ij}>q)$. 

A similar formulation has been used in \citep{VS10a} to recommend items to users in the ``who-bought-what'' 
network. Here, we show that this biased matrix factorization technique can be used to provably recover
$Y$. For convenience, we define the thresholding operator
$\text{thr}(x) = 1$ if $x>q$, and $\text{thr}(x)=0$ if $x\leq q$. 
We first define the recovery error as  $R(X) = \frac{1}{mn}\sum_{i,j} 1_{\text{thr}(X_{ij})\neq Y_{ij}}$, 
where $Y$ is the underlying 0-1 matrix. Define the label-dependent error:
\begin{equation}
  U_{\alpha}(x, a) = (1-\alpha)1_{\text{thr}(x)=1}1_{a=-1} + \alpha 1_{\text{thr}(x)=-1}1_{a=1}. 
\end{equation}
and $\alpha$-weighted expected error and expected $\alpha$-weighted loss:
\begin{equation}
  R_{\alpha, \rho}(X) = E\big[\sum_{i,j} U_{\alpha}(X_{ij}, A_{ij})\big], \  
  R_{l_\alpha, \rho}(X) = E\big[\sum_{i,j} l_\alpha (X_{ij}, A_{ij})\big]. 
\end{equation}
The following lemma is a special case of Theorem 9 in \citep{naga_nips}, 
showing that $R(X)$ and $R_{\alpha, \rho}(X)$
can be related by a linear transformation: 
\begin{lemma}
  For the choice $\alpha^*=\frac{1+\rho}{2}$ and $a=\frac{1+\rho}{2}$, 
  there exists a constant $b$ that is independent of $X$ such that, for any matrix $X$, 
  \begin{equation*}
    R_{\alpha^*, \rho}(X) = a R(X) + b. 
  \end{equation*}
  \label{thm:naga}
\end{lemma}
Therefore, minimizing the $\alpha$-weighed expected error in the partially observed situation 
is equivalent to minimizing the true recovery error $R$. By further relating $R_{\alpha^*, \rho}(X)$ and  $R_{\ell_{\alpha^*}, \rho}(X)$ we can show:
\begin{theorem}[Main Result 2]
  Let $\hat{X}$ be the minimizer of \eqref{eq:bias-factorization}, 
  and $\bar{X}$ be the thresholded 0-1 matrix of $\hat{X}$, then with probability at
  least $1-\delta$, we have 
  \begin{equation*}
    R(\bar{X}) \leq \frac{2\eta}{1+\rho}  \left(Ct\frac{\sqrt{n}+\sqrt{m}+\sqrt[4]{s}}{mn}
    + 3\frac{\sqrt{\log(2/\delta)}}{\sqrt{mn}(1-\rho)}\right), 
  \end{equation*}
  where $\eta = \max(1/q^2, 1/(1-q)^2) $ and  $C$ is a constant. 
  \label{thm:bias_mf_recover}
\end{theorem}
The average error is of the order of $O(\frac{1}{n(1-\rho)})$ when $M\in \R^{n\times n}$, where $1-\rho$ denotes the ratio of observed 1's, similar to the ShiftMC estimator. 
  
\section{PU Inductive Matrix Completion}
\label{sec:IMC}

In this section, we extend our approaches to inductive matrix completion problem, where in addition to the samples, row and column features
$F_u\in \R^{m\times d}, F_v \in \R^{n\times d}$ are also given. 
In the standard inductive matrix completion problem~\citep{dhillon_inductive}, the observations 
$A_{\Omega}$ are sampled from the groundtruth $M\in \R^{m\times n}$, 
and we want to recover $M$  by solving 
the following optimization problem:
\begin{equation}
  \min_{D\in \R^{d\times d}}  \sum_{i,j\in \Omega} (A_{ij}-(F_u D F_v^T)_{ij})^2 + \lambda \|D\|_*.  
  \label{eq:inductive}
\end{equation}
Matrix completion is a special case of inductive matrix completion when $F_u=I, F_v=I$. In the multi-label learning problem, $M$ represents the label matrix and $F_u$ corresponds to examples (typically $F_{v} = I$)~\citep{yu2014large,MX13a}.  This technique has also been applied to gene-disease prediction \citep{NN14a}, semi-supervised clustering \citep{JY13a}, and theoretically studied in \citep{dhillon_inductive}. 

The problem is fairly recent and we wish to extend PU learning analysis to this problem, which is also well motivated in many real world applications. For example, in multi-label learning
with partially observed labels, negative labels are usually not available. In the experiments, we will consider another interesting application --- semi-supervised clustering problem with only positive and unlabeled relationships. 

\subsection{Shifted Inductive Matrix Completion for Non-deterministic Setting }
\label{sec:SMC_inductive}
In the non-deterministic setting, we consider the inductive version
of ShiftMC: 
\begin{equation}
  \min_{D\in \R^{d\times d}} \sum_{i,j}\tilde{\ell}((F_u D F_v^T)_{ij}, A_{ij}) 
  \text{ such that }  \|D\|_*\leq t, \  1\geq F_u D F_v^T \geq 0,  
  \label{eq:inductive_shift}
\end{equation}
where the unbiased estimator of loss $\tilde{\ell}(\cdot)$ is defined in \eqref{eq:ltilde_define}. Note that we can assume that $F_u, F_v$  are orthogonal (otherwise
we can conduct a preprocessing step to normalize it). Let $\bu_i$ be the $i$-th row of $F_u$ (the feature for row $i$) and $\bv_j$ be the $j$-th row of $F_v$. Define constants $\X_u = \max_i \|\bu_i\|, \X_v = \max_j\|\bv_j\|$. 
\begin{equation*}
  \gamma = \min \left( \frac{\min_i \|\bu_i\|}{\X_u}, \frac{\min_i \|\bv_i\|}{\X_v}  \right).  
\end{equation*}
In practice if one does an instance-wise scaling of features, $\mu$ will be 1. 
Assume $F_u = U\Sigma V$, then we define $\bar{F}_u=U_{\bar{d}} \Sigma_{\bar{d}} V_{\bar{d}}$ where $\sigma_{\bar{d}}$ is the smallest singular value with $\sigma_d \geq \mu \sigma_1$. By the same way we can define $\bar{F}_v$. We assume the column space of the ground truth $M$ lies in $\text{span}(\bar{F}_u)$ and the row space of $M$ lies in $\text{span}(\bar{F}_v)$. We expect $\mu$ to be not too small, which indicates that the ground truth matrix lies in a more informative subspace of $F_u$ and $F_v$.  Since the output of inductive matrix completion is $F_u D F_v$, it can only recover the original matrix when the underlying matrix $M$ can be written in such form. Following \cite{MX13a,JY13a}, we assume the features are good enough such that $ M = F_u(F_u)^T M F_v (F_v)^T$. Recall $\|M\|_{*} \leq t$. We now extend Theorem \ref{thm:mc_random} to PU inductive matrix completion.
\begin{theorem}
  Assume $\hat{D}$ is the optimal solution of \eqref{eq:inductive_shift}  
  and the groundtruth $M$ is in the subspace formed by $F_u$ and $F_v$: $M = F_u (F_u)^T M F_v (F_v)^T$, 
  then 
  \begin{equation}
    \frac{1}{mn} \sum_{i,j} (M_{ij}-(F_{u}^{T}\hat{D}F_{v})_{ij})^2 \leq 6\frac{\sqrt{\log(2/\delta)}}{\sqrt{mn}(1-\rho)} 
   + \frac{4t\sqrt{\log 2d}}{\sqrt{mn}\sqrt{1-\rho}} \X_u \X_v . 
    \label{eq:inductive_shift_1}
  \end{equation}
  with probability at least $1-\delta$.
  \label{thm:inductive_shift_1}
\end{theorem}
Therefore if $t$ and $d$ are bounded, the mean square error of shiftMC is $O(1/n)$. 

\subsection{Biased Inductive Matrix Completion for Deterministic Setting}
\label{sec:BMC_inductive}
In the deterministic setting, we propose to solve the inductive version of BiasMC: 
\begin{equation}
  \hat{D} = \arg\min_{D: \|D\|_*\leq t} \alpha \sum_{i,j: A_{ij}=1} ((F_u D F_v^T)_{ij}-1)^2 +
  (1-\alpha)\sum_{i,j: A_{ij}=0} (F_u D F_v^T)_{ij}^2. 
  \label{eq:inductive_bias}
\end{equation}
The clean 0-1 matrix $Y$ can then be recovered by $\hat{Y}_{ij} = I((F_u \hat{D} F_v^T)_{ij} > q)$. 
\begin{equation}
  \hat{Y}_{ij} = \begin{cases}
    1 &\text{ if } (F_u \hat{D} F_v^T)_{ij} \geq q \\
    0 &\text{ if }  (F_u \hat{D} F_v^T)_{ij} < q. 
  \end{cases}
  \label{eq:inductive_thr}
\end{equation}
Similar to the case of matrix completion, Lemma \ref{thm:naga} shows that 
the expected 0-1 error $R(X)$ and the $\alpha$-weighted expected error in noisy observation $R_{\alpha, \rho}(X)$ can be related by a linear transformation 
when $\alpha^*=\frac{1+\rho}{2}$. With this choice of $\alpha^*$, Lemma \ref{thm:naga} continues to hold in this case, which allows us to extend Theorem \ref{thm:bias_mf_recover} to PU inductive matrix completion:
\begin{theorem}
  Let $\hat{D}$ be the minimizer of \eqref{eq:inductive_bias}
  with $\alpha^*=(1+\rho)/2$, and let $\hat{Y}$ be generated from $\hat{D}$ by thresholding, 
  then with probability at
  least $1-\delta$, we have 
  \begin{equation*}
    R(\hat{Y}) = \frac{1}{mn}\|Y-\hat{Y}\|^{2}_{F}  \leq \frac{2\eta}{1+\rho}  \left( \frac{4t\sqrt{\log 2d}}{\sqrt{mn}\sqrt{1-\rho}} \X_u \X_v
    + 6\frac{\sqrt{\log(2/\delta)}}{\sqrt{mn}(1-\rho)}\right), 
  \end{equation*}
  where $\eta = \max(1/q^2, 1/(1-q)^2) $.
  \label{thm:inductive_bias_recover}
\end{theorem}
Again, we have that if $t$ and $d$ are bounded, the mean square error of BiasMC is $O(1/n)$. 

\section{Optimization Techniques for PU Matrix Completion}
\label{sec:optimization}

In this section, we show that BiasMC can be solved very efficiently for large-scale (millions of rows and columns) datasets, 
and that ShiftMC can be solved efficiently after a relaxation. 

First, consider the optimization problem for BiasMC:
\begin{equation}
  \argmin_X \  \ \alpha \sum_{i,j: A_{ij}=1} (X_{ij}-1)^2 + (1-\alpha)\sum_{i,j: A_{ij}=0}
  X_{ij}^2 + \lambda \|X\|_* := f_b(X) + \lambda \|X\|_*, 
  \label{eq:biasMC_regularized}
\end{equation}
which is equivalent to the constrained problem \eqref{eq:bias-factorization}
with suitable $\lambda$. The typical proximal gradient descent update
is $X\leftarrow \S(X-\eta \nabla f_b(X), \lambda)$, where $\eta$ is 
the learning rate and $\S$ is the 
soft thresholding operator on singular values \citep{SJ09a}. 
The (approximate) SVD of $G:=(X-\eta\nabla f_b(X))$ can be computed efficiently
using power method or Lanczos algorithm 
if we have a fast procedure to compute $GP$ for a tall-and-thin
matrix $P\in \R^{n\times k}$. 
In order to do so, we first rewrite $f_b(X)$ as
\begin{equation}
  f_b(X) = (1-\alpha)\|X-A\|_F^2 + (2\alpha-1)\sum_{i,j: A_{ij}=1} (X_{ij}-A_{ij})^2.
  \label{eq:trick}
\end{equation}
Assume the current solution is stored in a low-rank form $X=WH^T$ and $R = (X-A)_{\Omega_1}$ is the 
residual on $\Omega_1$, 
then 
\begin{align*}
  GP &= X P - 2\eta\left( (1-\alpha)(X-A)+ (2\alpha-1)(X-A)_{\Omega_1} \right) P \\ 
  &= (1-2\eta(1-\alpha))WH^T P + 2\eta(1-\alpha)AP - 2\eta (2\alpha-1)R P, 
\end{align*}
where the first term can be computed in $O(mk^2+nk^2)$ flops, and the remaining
terms can be computed in $O(|\Omega_1|k)$ flops. With this approach, we can efficiently compute the proximal operator. This can also be applied to other faster nuclear norm solvers (for example, \citep{CJH13c}). 

Next we show that the non-convex form of BiasMC can also be efficiently solved, 
and thus can scale to millions of nodes and billions of observations. 
It is well known that the nuclear norm regularized problem $\min_X f_b(X)+\lambda\|X\|_*$ 
is equivalent to 
\begin{equation}
  \min_{W\in \R^{m\times k},H\in \R^{n\times k}} f_b(WH^T) +\frac{\lambda}{2}(\|W\|_F^2 + \|H\|_F^2)  
  \label{eq:fb_nonconvex}
\end{equation}
when $k$ is sufficiently large. We can use a trick similar to \eqref{eq:trick} to compute the gradient and Hessian efficiently: 
  \begin{align*}
  \nabla_{W} f_b(WH^T) &= 2(1-\alpha) (WH^T H -AH) + 2(2\alpha-1) R_{\Omega} H \\
  \text{ and }\nabla^2_{W_{i,\cdot}} f_b(WH^T) &= 
  2(1-\alpha) H^T H + 2(2\alpha-1) H_{\Omega_i}^T H_{\Omega_i}, 
\end{align*}
where $H_{\Omega_i}$ is the sub-matrix with columns $\{\bh_j: j\in \Omega_i\}$, 
and $\Omega_i$ is the column indices of observations in the $i$-th row. 
Thus, we can efficiently apply Alternating Least Squares (ALS) or Coordinate Descent (CD) for solving \eqref{eq:fb_nonconvex}.  For example, when applying CCD++ in \citep{PMF}, each coordinate descent update
only needs $O(|\Omega_i|+k)$ flops. We apply this technique to solve large-scale
link prediction problems~(see Section \ref{sec:experiments}). 

The optimization problem for ShiftMC is harder to solve because of the 
bounded constraint. We can apply the bounded matrix factorization technique 
\citep{BMF} to solve the non-convex form of \eqref{eq:alg2}, where
the time complexity is $O(mn)$ because of the constraint
$0\leq (WH^T)_{ij}\leq 1$ for all $(i,j)$. To scale it to large datasets, we relax the bounded constraint
and solve:
\begin{equation}
  \min_{W\in \R^{m\times k}, H \in \R^{n\times k}} \|A-WH^T\|_F^2 + \frac{\lambda}{2}(\|W\|_F^2 + \|H\|_F^2) \text{ s.t. } 0\leq W, H\leq \sqrt{1/k}
  \label{eq:shiftMC_relax}
\end{equation}
This approach (ShiftMC-relax) is easy to solve by ALS or CD with $O(|\Omega|k)$ complexity per sweep (similar to the BiasMC). In our experiments, we show ShiftMC-relax performs even better than shiftMC in practice. 

\section{Experiments}
\label{sec:experiments}
We first use synthetic data to show that our bounds are meaningful and then demonstrate the effectiveness of our algorithms in real world applications. 
\begin{figure}[t!h]
  \begin{tabular}{cc}
      \subfigure[Synthetic data: non-deterministic setting. ] {
    \includegraphics[width=0.45\textwidth]{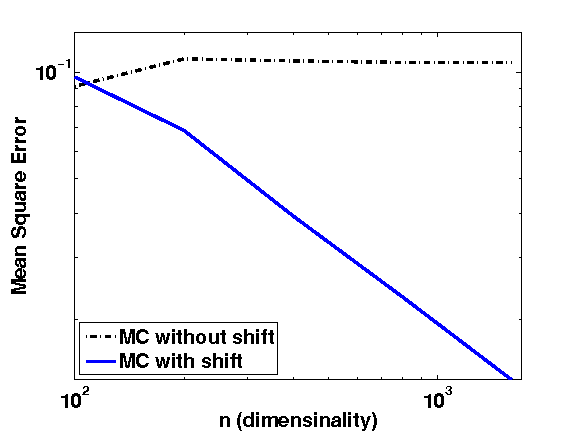}
    } &
    \subfigure[Synthetic data: deterministic setting.] {
    \includegraphics[width=0.45\textwidth]{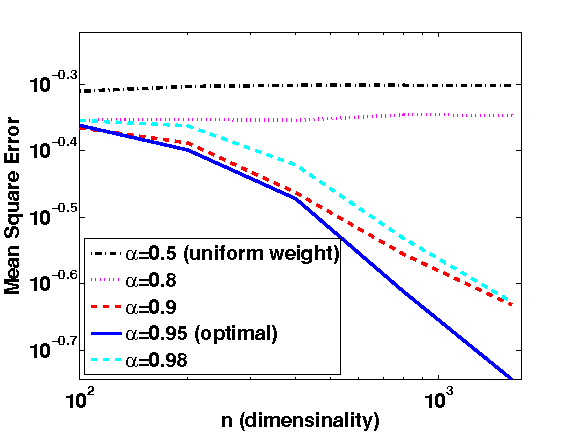}
    } \\ 
    \subfigure[FPR-FNR on \GrQc dataset. ] {
    \includegraphics[width=0.45\textwidth]{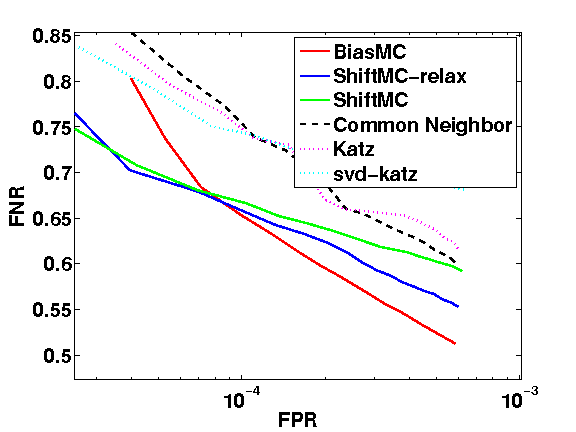}
    }  & 
    \subfigure[FPR-FNR on \HepPh dataset. ] {
    \includegraphics[width=0.45\textwidth]{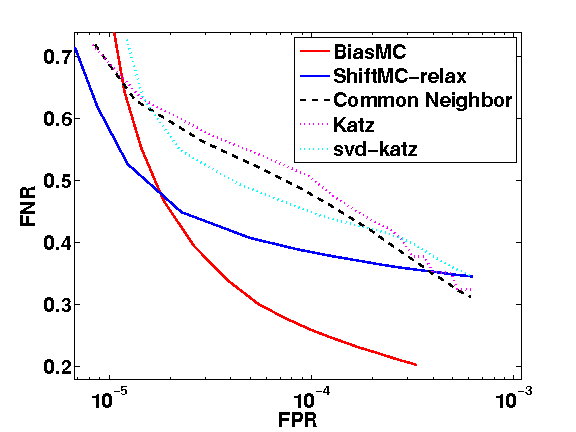}
    }\\ 
    \subfigure[FPR-FNR on \LiveJournal dataset. ] {
    \includegraphics[width=0.45\textwidth]{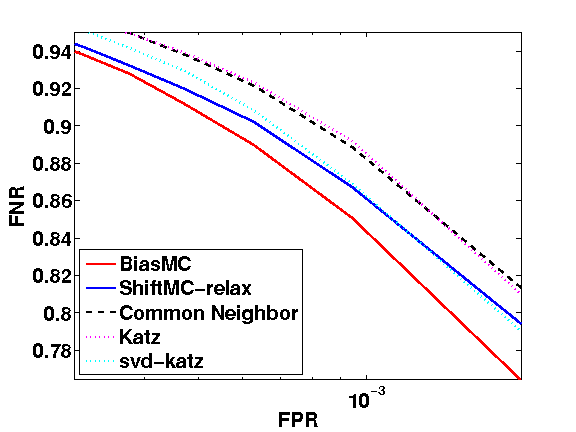}
    } &
    \subfigure[FPR-FNR on \MySpace dataset. ] {
    \includegraphics[width=0.45\textwidth]{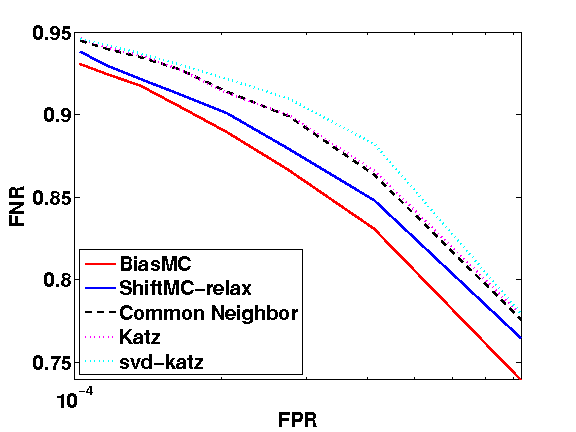}
    } 
  \end{tabular}
  \caption{ (a)-(b): Recovery error of ShiftMC and BiasMC on synthetic data.
   We observe that without shifting or biasing, error does not decrease with $n$ (the black lines).  The error of our estimators decreases approximately as $\frac{1}{n}$, as proved in Theorems \ref{thm:mc_random} and \ref{thm:bias_mf_recover}.  (c)-(f): Comparison of link prediction methods. ShiftMC and BiasMC consistently perform better than the rest.
 \label{fig:linkprediction} }
\end{figure}

\subsection{Synthetic Data}
We assume the underlying matrix $M\in \R^{n\times n}$ is generated by $UU^T$, 
where $U\in \R^{n\times k}$ is the orthogonal basis of a random Gaussian $n$ by $k$ matrix
with mean 0 and variance 1. For the non-deterministic setting, we linearly scale
$M$ to have values in $[0,1]$, and then generate training samples as described Section \ref{sec:settings}. For deterministic setting,  we choose $q$ so that $Y$ has equal number of zeros and ones. We fix $\rho=0.9$ (so that only 10\% 1's are observed). From Lemma \ref{thm:naga}, $\alpha=0.95$ is optimal. We fix $k=10$, and test our algorithms with different sizes $n$. The results are shown in Figure \ref{fig:linkprediction}(a)-(b). 
Interestingly, the results reflect our theory: error of our estimators decreases with $n$; in particular, error linearly decays with $n$ in log-log scaled plots, which suggests a rate of $O(1/n)$, as shown in Theorems \ref{thm:mc_random} and \ref{thm:bias_mf_recover}. Directly minimizing $\|A-X\|_F^2$ gives very poor results. For BiasMF, we also plot the performance of estimators with various $\alpha$ values in Figure \ref{fig:linkprediction}(b).  As our theory suggests, $\alpha=\frac{1+\rho}{2}$ performs the best. We also observe that the error is well-behaved in a certain range of $\alpha$. A principled way of selecting $\alpha$ is an interesting problem for further research. 

\subsection{Parameter Selection}
Before showing the experimental results on real-world problems, 
we discuss the selection of the parameter $\rho$ in our PU matrix completion 
model (see eq~\eqref{eq:random_A}). 
Note that $\rho$ indicates the noise rate of flipping a 1 to 0.  
If there are equal number of positive and negative elements in the 
underlying matrix $Y$, we will have $\rho=1-2s$ 
where $s=(\text{\# positive entries})/(\text{\# total entries})$. 
In practice (e.g., link prediction problems) number of 1's are usually
less than number of 0 in the underlying matrix, but we do not know the ratio. 
Therefore, in all the experiments 
we chose $\rho$ from the set $\{1-2s, 10(1-2s), 100(1-2s), 1000(1-2s)\}$
based on a random validation set, and use the corresponding $\alpha$
in the optimization problems. 

\subsection{Matrix completion for link prediction}
One of the important applications that motivated our analysis in this paper is the link prediction problem. Here, we are given $n$ nodes (users) and a set of edges $\Omega_{train}$ (relationships) and the goal is to predict \emph{missing} edges, i.e. $\Omega_{test}$. We use 4 real-world datasets: 2 co-author networks \GrQc (4,158 nodes and 26,850 edges) and \HepPh (11,204 nodes and 235,368 edges), where we randomly split edges into training and test such that $|\Omega_{train}| = |\Omega_{test}|$;  2 social networks \LiveJournal (1,770,961 nodes, $|\Omega_{train}|$ = 83,663,478 and $|\Omega_{test}|$ = 
2,055,288) and \MySpace (2,137,264 nodes, $|\Omega_{train}|$ = 90,333,122 and $|\Omega_{test}|$ = 1,315,594), where train/test split is done using timestamps. 
For our proposed methods BiasMC, 
ShiftMC and ShiftMC-relax, we solve the non-convex form with $k=50$ for \GrQc, 
\HepPh and $k=100$ for \LiveJournal and \MySpace.  The $\alpha$ and $\lambda$ 
values are chosen by a validation set.  

We compare with competing link 
prediction methods \citep{linkprediction} Common Neighbors, Katz, and SVD-Katz 
(compute Katz using the rank-$k$ approximation, $A\approx U_k\Sigma_k V_k$). 
Note that the classical matrix factorization approach in this case is 
equivalent to SVD on the given 0-1 training matrix, and 
SVD-Katz slightly improves over SVD by further computing the Katz
values based on the low rank approximation (see~\cite{linkprediction}), 
so we omit the SVD results in the figures. 

Based on the training matrix, each link prediction method will output a list 
of $k$ candidate entries. The quality of the top-$k$ entries can be evaluated 
by computing the False Positive Rate (FPR) and False Negative Rate (FNR) 
defined by 
\begin{equation*}
  \text{FPR} = \frac{\text{\# of incorrectly predicted links}}{\text{\# of non-friend links}}, \  
  \text{FNR} = 1 - \frac{\text{\# of correctly predicted links}}{\text{\# of actual links}},
\end{equation*}
where the groundtruth is given in the test snapshot. 
The results are shown in Figure \ref{fig:linkprediction}. 
\GrQc is a small dataset, so we can solve the original
ShiftMC problem accurately, although ShiftMC-relax achieves a similar performance here. For larger datasets, we show only the performance of ShiftMC-relax.  In general BiasMC performs the best, and ShiftMC tends to perform better in the beginning. 
Overall, our methods achieve lower FPR and FNR comparing to other methods, 
which indicate that we obtain a better link prediction model
by solving the PU matrix completion problem. 
Also, BiasMC is highly efficient --- it takes 516 seconds for 10 coordinate descent sweeps on the largest dataset (\MySpace),  whereas computing top 100 eigenvectors using \texttt{eigs} in \textsc{Matlab} requires 2408 seconds.

\begin{figure}[t]\centering
  \begin{tabular}{cc}
    \subfigure[Mushroom dataset.  ] {
    \includegraphics[width=0.45\textwidth]{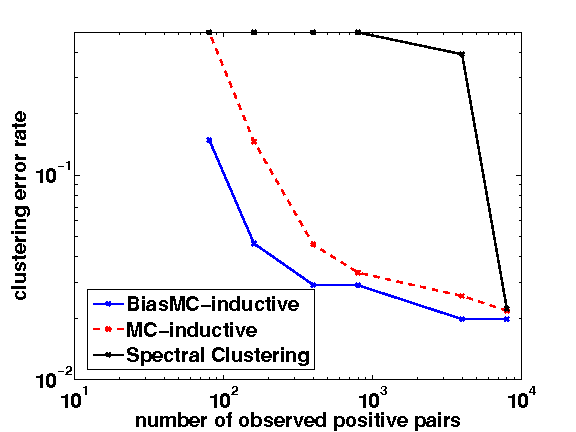}
    } &
    \subfigure[Segment dataset.  ] {
    \includegraphics[width=0.45\textwidth]{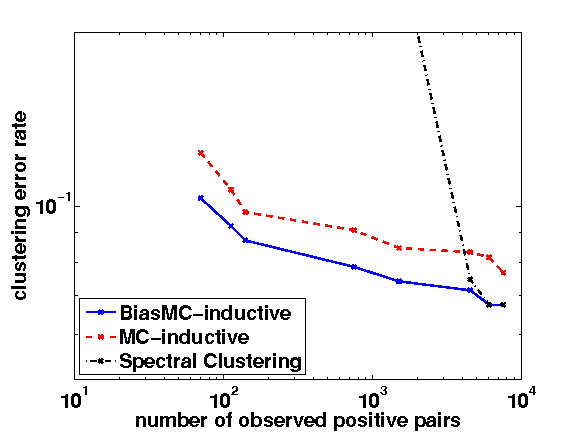}
    }  
  \end{tabular}
   \caption{ Semi-supervised clustering performance of BiasMC-inductive on two real datasets. BiasMC-inductive performs better than MC-inductive (treats unlabeled relationships as zeros) and spectral clustering (does not use features). BiasMC-inductive achieves under 10\% error using just 100 samples. \label{fig:inductive}}
\end{figure}

\subsection{Inductive matrix completion}
We use the semi-supervised clustering problem to evaluate our PU inductive matrix completion methods. 
PU inductive matrix completion can be applied to many real-world problems, 
including recommender systems with features and 0-1 observations, 
and the semi-supervised clustering problem when we can only observed
positive relationships. Here we use the latter as an example to demonstrate
the usefulness of our algorithm. 

In semi-supervised clustering problems, we are given $n$ samples with features 
$\{\bx_i\}_{i=1}^n$ and pairwise relationships $A\in \R^{n\times n}$, 
where 
$A_{ij}=1$ if two samples are in the same cluster, $A_{ij}=-1$ if they are in 
different clusters, and $A_{ij}=0$ if the relationship is unobserved. Note 
that the groundtruth matrix $M \in \{+1,-1\}^{n\times n}$ exhibits a simple 
structure and is a low rank as well as low trace norm matrix; it is shown in 
\citep{JY13a} that we can recover $M$ using IMC when there are both positive 
and negative observations.  

Now we consider the setting where only positive relationships are observed, so 
$A$  is a 0-1 matrix. 
We show that biased IMC can recover $M$ using 
very few positive relationships. We test the algorithms on two datasets: the 
Mushroom dataset with $8142$ samples, $112$ features, and 2 classes; the 
Segment dataset with $2310$ samples, $19$ features, and $7$ classes.  The 
results are presented in Figure \ref{fig:inductive}.

We compare 
BiasMC-inductive with (a) MC-inductive, which considers all the unlabeled 
pairs as zeros and minimizes $\|F_u D F_v^T - A\|_F^2$, and (b) spectral 
clustering, which does not use feature information. 
Since the data is from classification datasets, the ground truth $M$
is known and can be used to evaluate the results. 
In Figure \ref{fig:inductive}, the vertical axis is the
clustering error rate defined by 
\begin{equation*}
  \text{(number of entries in $M$ predicted with correct sign)}/n^2. 
\end{equation*}
Figure 
\ref{fig:inductive} shows that BiasMC-inductive is much better than other 
approaches for this task.

\section{Conclusions}
\label{sec:conclusions}
Motivated by modern applications of matrix completion, our work attempts to bridge the gap between the theory of matrix completion and practice. We have shown that even when there is noise in the form of one-bit quantization as well as one-sided sampling process revealing the measurements, the underlying matrix can be accurately recovered. We have considered two recovery settings, both of which are natural for PU learning, and have provided similar recovery guarantees for the two. Our error bounds are strong and useful in practice. Our work serves to provide the first theoretical insight into the biased matrix completion approach that has been employed as a heuristic for similar problems in the past. Experimental results on synthetic data conform to our theory; effectiveness of our methods are evident for the link prediction task in real-world networks. A principled way of selecting or estimating the bias $\alpha$ in BiasMC seems worthy of exploration given our encouraging results.

\bibliographystyle{plainnat}
\bibliography{oneclass}

\setcounter{section}{0}
\renewcommand\thesection{\Alph{section}}
\section{Proofs}
\label{sec:proofs}

\subsection{Proof of Lemma \ref{thm:2}}
\label{app:thm:2}
\begin{proof}
  \begin{align*}
  \frac{1}{mn}  E \bigg[\sum_{i,j} \tilde{l}(X_{ij}, A_{ij})\bigg] &= 
    \frac{1}{mn}\sum_{i,j}E \bigg[\tilde{l}(X_{ij}, A_{ij})\bigg] \\
    &= \frac{1}{mn} \sum_{i,j} \left( P(Y_{ij}=0) X_{ij}^2 + P(Y_{ij}=1) \bigg(\rho X_{ij}^2 
    + (1-\rho)(\frac{(X_{ij}-1)^2 - \rho X_{ij}^2}{1-\rho})\bigg)\right) \\
    &= \frac{1}{mn} \sum_{i,j} \left( P(Y_{ij}=0) X_{ij}^2 + P(Y_{ij}=1)(X_{ij}-1)^2 \right) \\
    &= \frac{1}{mn} E \bigg[ \sum_{i,j}(X_{ij}-Y_{ij})^2 \bigg] = \frac{1}{mn} E\bigg[ l(X_{ij}, Y_{ij}) \bigg]. 
  \end{align*}
\end{proof}
\subsection{Proof of Theorem \ref{thm:mc_empirical_diff}}
\label{app:thm:mc_empirical_diff}
\begin{proof}
  We want to bound $\sup_{X\in \X} \Big|\hat{R}_{\tilde{l}}(X) - E_A [R_{\tilde{l}}(X)]\Big|$.  
First, 
\begin{equation*}
  E_A [R_{\tilde{l}}(X)] \leq \hat{R}_{\tilde{l}} (X) + \sup_{X\in \X} \Big(
  E_A[\frac{1}{mn}\sum_{i,j} \tilde{l}(X_{ij}, A_{ij})] - \frac{1}{mn}\sum_{i,j}\tilde{l}
  (X_{ij}, A_{ij})\Big). 
\end{equation*}
Apply McDiarmid's Theorem in \citep{naga_book}; 
since each $\tilde{l}(X_{ij}, A_{ij})$ can be either $X_{ij}^2$ or $\frac{(X_{ij}-1)^2 - \rho X_{ij}^2}{1-\rho}$, 
when changing one random variable $A_{ij}$, 
in the worst case the quantity 
$\sup_{X\in \X} \Big( E_A[R_{\tilde{l}}(X)]-\frac{1}{mn}\sum_{i,j}\tilde{l}(X_{ij}, A_{ij})\Big)$
can be changed by 
\begin{equation*}
  \Big| X_{ij}^2 - \frac{(X_{ij}-1)^2 - \rho X_{ij}^2}{1-\rho}\Big| \leq  \Big|\frac{2X_{ij}+1}{1-\rho}\Big| \leq 
  \frac{3}{1-\rho}. 
\end{equation*}
So by McDiarmid's Theorem, with probability $1-\delta/2$, 
\begin{equation}
  \sup_{X\in\X} \Big( E_A(R_{\tilde{l}}(X)) - \frac{1}{mn} \sum_{i,j}\tilde{l}(X_{ij}, A_{ij}) \Big)
  \leq E_A\left[  \sup_{X\in \X} \big(E_A(R_{\tilde{l}}(X))-\frac{1}{mn}\sum_{i,j}\tilde{l}(X_{ij}, A_{ij})  \big)\right] + 3\frac{\sqrt{\log(2/\delta)}}{\sqrt{mn}(1-\rho)}. 
\end{equation}

Also, 
\begin{align}
  &E_A\left[ \sup_{X\in \X}\Big(E_A[R_{\tilde{l}}(X)] -\frac{1}{mn}\sum_{i,j}\tilde{l}(X_{ij}, A_{ij}) \Big) \right] \\
  &\leq E_{A, \tilde{A}} \left[ \sup_{X\in \X} \Big( \frac{1}{mn} \sum_{i,j}\tilde{l}(X_{ij}, \tilde{A}_{ij})
  -\frac{1}{mn} \sum_{i,j} \tilde{l}(X_{ij}, A_{ij})\Big) \right] \\
  &= E_{A,\tilde{A}} \left[ \sup_{X\in\X} \frac{1}{mn} \sum_{i,j}(\tilde{l}(X_{ij}, \tilde{A}_{ij})-\tilde{l}
  (X_{ij}, A_{ij})) \right] \label{eq:abc11}\\
  &= \frac{1}{mn} E_{A, A', \sigma} \left[ \sup_{X\in \X} \sum_{i,j: M_{ij}=1} \sigma_{ij}
  (\tilde{l}(X_{ij}, \tilde{A}_{ij})-\tilde{l}(X_{ij}, A_{ij}))  \right] \label{eq:abc22} \\
  &\leq \frac{1}{mn} E_{A, \sigma} \left[ \sup_{X\in \X} \sum_{i,j: M_{ij}=1} \sigma_{ij}
  \tilde{l}(X_{ij}, A_{ij})\right] \label{eq:bound1}
\end{align}
where $\sigma_{ij}$ are random variables with half chance to be +1
and half chance to be -1. 
Where from \eqref{eq:abc11} to \eqref{eq:abc22} we use the fact that $A_{ij}=0$ with probability 1 if $M_{ij}=0$. 
Next we want to bound the Rademacher complexity $E_{A, \sigma} \left[ \sup_{X\in \X} \sum_{i,j: M_{ij=1}} \sigma_{ij}\tilde{l}(X_{ij}, A_{ij})\right]$. 
When $M_{ij}=1$, 
\begin{equation*}
  \tilde{l}(X_{ij}, A_{ij}) = \begin{cases}
    X_{ij}^2 &\text{ with probability } \rho \\
    \frac{(X_{ij}-1)^2 - \rho X_{ij}^2}{1-\rho} &\text{ with probability } 1-\rho 
  \end{cases}
\end{equation*}
Since $0\leq X_{ij}\leq 1$, the Lipschitz constant for $\tilde{l}(X_{ij}, A_{ij})$
is at most $1/(1-\rho)$, so
\begin{align*}
  \eqref{eq:bound1} &\leq \frac{1}{mn} E_{\sigma} (\sup_{X\in \X} \sum_{i,j: M_{ij}=1}\sigma_{ij}X_{ij}) \\
  &\leq \frac{1}{(1-\rho)mn} E_{\sigma} \left[\sup_{X\in \X} \|\Proj_{M_{ij}=1}(\sigma) \|_2 \|X\|_*\right]. 
\end{align*}
As pointed out in \cite{OS11a}, we can then apply the main Theorem in \cite{Latala05}, when $Z$ is an independent zero mean random matrix, 
\begin{equation*}
  E[\|Z\|_2] \leq C\left( \max_i \sqrt{\sum_{j}E[Z_{ij}^2]}+\max_j \sqrt{\sum_i E[Z_{ij}^2]} + \sqrt[4]{\sum_{ij}E[Z_{ij}^4]} \right)
\end{equation*}
with a universal constant $C$. 

So in our case $E[\|\sigma\|_2] \leq C\left( \sqrt{n} +\sqrt{m} + \sqrt[4]{s} \right)$, 
so $\eqref{eq:bound1} \leq tC\frac{\sqrt{n}+\sqrt{m}+\sqrt[4]{s}}{(1-\rho)mn}$. 
\end{proof}

\subsection{Proof of Theorem \ref{thm:mc_random}}
\label{app:thm:mc_random}

Let $\hat{X}$ be the minimizer of \eqref{eq:alg1}, and 
\begin{equation*}
  P := tC\frac{\sqrt{n}+\sqrt{m}+\sqrt[4]{s}}{(1-\rho)mn}
    + 3\frac{\sqrt{\log(2/\delta)}}{\sqrt{mn}(1-\rho)}, 
\end{equation*}
we have
\begin{align*}
  E\big[ \frac{1}{mn}\sum_{i,j} (\hat{X}_{ij}-Y_{ij})^2 \big] &=   E\big[\frac{1}{mn} \sum_{i,j} \tilde{l}(\hat{X}_{ij}, A_{ij}) \big] \ \ \ (\text{Lemma \ref{thm:2}}) \\
  &\leq \hat{R}_{\tilde{l}}(\hat{X}) + P \ \ \ (\text{Theorem \ref{thm:mc_empirical_diff}})\\
  &\leq \hat{R}_{\tilde{l}}(M) + P \ \ \ (\text{by the definition of $\hat{X}$}) \\
  &\leq E\big[ \frac{1}{mn} \tilde{l}(M_{ij}, A_{ij}) \big] + 2P \ \ \ (\text{Theorem \ref{thm:mc_empirical_diff}}) \\
  &= E\big[ \frac{1}{mn} \sum_{i,j}(M_{ij}-Y_{ij})^2 \big] \ \ \ \text{(Lemma \ref{thm:2})} 
\end{align*}
Therefore 
\begin{equation*}
  \frac{1}{mn} \sum_{i,j} E\big[ (X_{ij}-Y_{ij})^2 - (M_{ij}-Y_{ij})^2 \big] \leq 2P. 
\end{equation*}
Since $P(Y_{ij}=1) = M_{ij}$,  we have
\begin{align*}
  E\big[ (X_{ij}-Y_{ij})^2 - (M_{ij} - Y_{ij})^2 ] 
  &= M_{i,j}\big( (X_{ij}-1)^2 - (M_{ij}-1)^2 \big) + (1-M_{ij})(X^2_{ij} - M^2_{ij}) \\
  &=  (X_{ij}-M_{ij})^2, 
\end{align*}
therefore
\begin{equation*}
  \frac{1}{mn} \sum_{i,j}(X_{ij}-M_{ij})^2 \leq 2P. 
\end{equation*}

\subsection{Proof of Theorem \ref{thm:bias_mf_recover}}
\label{app:thm:bias_mf_recover}
\begin{proof}
We want to show 
\begin{equation}
  R_{\alpha, \rho}(X) - \min_X R_{\alpha, \rho}(X) \leq \eta 
  (R_{l_\alpha, \rho}(X) - \min_{X} R_{l_\alpha, \rho}(X)), 
  \label{eq:Rbound1}
\end{equation}
where $\eta = \max(1/q^2, 1/(1-q)^2)$. 
Consider the following two cases. If $Y_{ij}=0$, 
then 
\begin{align*}
  & R_{\alpha, \rho} (X_{ij}) = \alpha 1_{X_{ij}>q}, \ \ \min_{X_{ij}}R_{\alpha, \rho}(X_{ij})=0 \\
  & R_{l_{\alpha}, \rho}(X_{ij}) = \alpha X_{ij}^2, \ \ \min_{X_{ij}}R_{l_{\alpha}, \rho}(X_{ij}) = 0, 
\end{align*}
so the left hand side of \eqref{eq:Rbound1} is $\alpha 1_{X_{ij}>q}$ and the right hand side is 
$\alpha X_{ij}^2$. Therefore we can simply verify that \eqref{eq:Rbound1} holds with $\eta = 1/q^2$. 
For the second case if $Y_{ij}=1$, 
\begin{align*}
  & R_{\alpha, \rho} (X_{ij}) = \rho(1-\alpha^*)1_{X_{ij}>q} + (1-\rho)\alpha^* 1_{X_{ij}<q} = \frac{(1-\rho)(1+\rho)}{2}1_{X_{ij}<q} + \frac{\rho(1-\rho)}{2} 1_{X_{ij}>q},  \\
  & R_{l_{\alpha}, \rho}(X_{ij}) = \frac{(1-\rho)(1+\rho)}{2}(X_{ij}-1)^2 + \frac{\rho(1-\rho)}{2}X_{ij}^2.  
\end{align*}
We can see $(1-\rho)(1+\rho)1_{X_{ij}<q} \leq \frac{1}{(1-q)^2}(1-\rho)(1+\rho)(X_{ij}-1)^2$ and
$\rho(1-\rho)1_{X_{ij}<q}<\frac{1}{q^2}\rho(1-\rho)X_{ij}^2$, so both will satisfied by our chosen $\eta$. 

Next we compute $\min_{X_{ij}}R_{\alpha, \rho}(X_{ij})$ and $\min_{X_{ij}}R_{l_\alpha, \rho}(X_{ij})$. 
By definition 
\begin{equation*}
  R_{\alpha^*, \rho}(X_{ij}) = \begin{cases}
    \rho(1-\alpha^*) = \frac{\rho(1-\rho)}{2} &\text{ if } X_{ij}>q \\
    (1-\rho)\alpha^* = \frac{(1+\rho)(1-\rho)}{2} & \text{ if } X_{ij}<q. 
  \end{cases}
\end{equation*}
Therefore 
\begin{equation}
  \min_{X_{ij}}R_{\alpha, \rho}(X_{ij}) = \frac{\rho(1-\rho)}{2}. 
  \label{eq:ralpha_min}
\end{equation}
On the other hand, 
\begin{equation*}
  R_{l_\alpha, \rho}(x) = \rho(1-\alpha^*) x^2 + (1-\rho)\alpha^* (x-1)^2
  = \frac{\rho(1-\rho)}{2}x^2 + \frac{(1-\rho)(1+\rho)}{2} (x-1)^2. 
\end{equation*}
Taking gradient equals to zero we get $x^* = \frac{\rho+1}{2\rho+1}$, 
and therefore
\begin{equation}
  \min_x R_{l_\alpha, \rho}(x) =  \frac{(1+\rho)\rho(1-\rho)}{2} 
  \leq \rho(1-\rho). 
  \label{eq:rlalpha_min}
\end{equation}
Combining \eqref{eq:ralpha_min} and \eqref{eq:rlalpha_min}, we have
\begin{equation*}
  \min_x R_{\alpha, \rho} \geq \min_x R_{l_\alpha, \rho}/2, 
\end{equation*}
therefore we need $\eta \geq 2$. But this is satisfied by $\eta = \max(1/q^{2}, 1/(1-q)^{2})$.
Combining the above arguments, we proved that 
  \eqref{eq:Rbound1} holds. 

  Next we show an upper bound of \eqref{eq:Rbound1}. Using the proof similar to Theorem  
  \ref{thm:mc_empirical_diff} we have
  \begin{equation}
R_{l_\alpha, \rho}(X) - \min_{X} R_{l_\alpha, \rho}(X) \leq
Ct\frac{\sqrt{n}+\sqrt{m}+\sqrt[4]{s}}{mn}
+ 3\frac{\sqrt{\log(2/\delta)}}{\sqrt{mn}(1-\rho)}
    \label{eq:Rlbb}
  \end{equation}

  Now for the left hand side $R_{\alpha, \rho}(X)-\min_X R_{\alpha, \rho}(X)$. By Theorem 
  \ref{thm:naga}, we know that
  \begin{equation}
    R_{\alpha^*, \rho}(X) - \min_X R_{\alpha^*, \rho}(X) 
    = \bigg(\frac{1+\rho}{2}\bigg)R(X). 
    \label{eq:Ralphabb}
  \end{equation}
  Here we use the fact that $\min_X R(X)=0$ because $R(Y)=0$, and the term $B$ vanished
  because it is a constant for both sides. Combining \eqref{eq:Ralphabb}, \eqref{eq:Rlbb}
  and  \eqref{eq:Rbound1}, we have
  \begin{equation*}
    \bigg(\frac{1+\rho}{2}\bigg) R(X) \leq \eta \left(Ct\frac{\sqrt{n}+\sqrt{m}+\sqrt[4]{s}}{mn}
    + 3\frac{\sqrt{\log(2/\delta)}}{\sqrt{mn}(1-\rho)}\right), 
  \end{equation*}
  therefore
  \begin{equation*}
    R(X) \leq \frac{2\eta}{1+\rho}  \left(Ct\frac{\sqrt{n}+\sqrt{m}+\sqrt[4]{3s}}{mn}
    + 3\frac{\sqrt{\log(2/\delta)}}{\sqrt{mn}(1-\rho)}\right). 
  \end{equation*}

\end{proof}

\subsection{Proof of Theorem \ref{thm:inductive_shift_1}}
\label{app:thm:inductive_shift_1}

\begin{proof}
  For convenience, we let $X=F_u D F_v^T$. We first apply the same
  argument in to the proof in Appendix \ref{app:thm:mc_empirical_diff} to get 
  \eqref{eq:bound1}. Now we want to bound the Rademacher compelxity 
  $E_{A, \sigma}[\sup_{w\in \W} \sum_{i,j} \sigma_{ij}\tilde{l}(X_{ij}, A_{ij})]$ (upper bound
  of \eqref{eq:bound1}). 
  Since $\tilde{l}(X_{ij}, A_{ij})$ is Lipchitz continuous with constant $\frac{1}{1-\rho}$
  (use the fact that $X_{ij}$ is bounded between 0 and 1), 
  we have $\tilde{l}(X_{ij}, A_{ij})\leq \frac{1}{1-\rho}(X_{ij}-A_{ij})$. 
  Therefore, 
  \begin{align*}
    &E_{A, \sigma}[\sup_{w\in\W} \sum_{i,j: A_{ij}=1} \sigma_{ij}\tilde{l}((F_u D F_v^T)_{ij}, 
    A_{ij})]  \\ 
    &\leq  E_{A, \sigma}[\sup_{w\in \W} \sum_{i,j: A_{ij}=1} \frac{\sigma_{ij}}{1-\rho}(F_u D F_v^T)_{ij} ] + E_{A, \sigma}[\frac{\sigma_{ij}}{1-\rho}] \\
    &= \frac{1}{1-\rho} E_{A,\sigma}[\sup_{w\in \W} \sum_{i,j: A_{ij}=1} \sigma_{ij} \text{tr}(\bu_i \bv_j^T D)  ] 
  \end{align*}

  We then use the following Lemma, which is a special case of Theorem 1 in \cite{SK07a}
  when taking $\|\cdot\|$ to be the matrix 2 norm and $\|\cdot\|_*$ (the dual norm)
  is the trace norm:
  \begin{lemma}
    Let $\D := \{D \mid D\in \R^{d\times d} \text{ and } \|D\|_* \leq \D_1\}$ (where $\|W\|_*$ is the 
    trace norm of $W$), and $\W = \max_i \|W_i\|_2$, then 
    \begin{equation*}
      E_{\sigma} [\sup_{D\in \D} \frac{1}{p} \sum_{i=1}^m \sigma_i \text{tr}(DW_i)] \leq 
      2\W\D_1 \sqrt{\frac{\log 2d}{p}}. 
    \end{equation*}
  \end{lemma}

  Now the set $\D$ is $\{D: \|D\|_*\leq t\}$ and number of terms that $A_{ij}=1$ is
  $p= n^2(1-\rho)$, 
  so using the above lemma we have
  \begin{align*}
E_{A, \sigma}[\sup_{D\in\D} \sum_{i,j: A_{ij}=1} \sigma_{ij}\tilde{l}((F_u D F_v^T)_{ij}, A_{ij})]
&\leq \frac{2}{1-\rho} t \left( \max_{i,j} \|\bu_i \bv_j^T\|_2  \right) \sqrt{\log 2d} \sqrt{p}  \\
&\leq \frac{2}{1-\rho} t \X_u \X_v \sqrt{\log 2d} \sqrt{mn} \sqrt{1-\rho} \\
&= \frac{2\sqrt{mn} t\sqrt{\log 2d}}{\sqrt{1-\rho}} \X_u \X_v. 
  \end{align*}
  Therefore, 
  \begin{equation*}
    \frac{1}{mn} E_{A, \sigma}[\sup_{D\in\D} \sum_{i,j: A_{ij}=1} \sigma_{ij}\tilde{l}((F_u D F_v^T)_{ij}, A_{ij})]
    \leq \frac{2t\sqrt{\log 2d}}{\sqrt{mn}\sqrt{1-\rho}} \X_u \X_v. 
  \end{equation*}
  Combined with other part of the proof of Theorem \ref{thm:mc_empirical_diff} we have
  \begin{equation*}
    \frac{1}{mn} \sum_{i,j} (M_{ij}-F_u \hat{D} F_v^T)^2 \leq 6\frac{\sqrt{\log(2/\delta)}}{\sqrt{mn}(1-\rho)} 
    + \frac{4t\sqrt{\log 2d}}{\sqrt{mn}\sqrt{1-\rho}} \X_u \X_v . 
  \end{equation*}
\end{proof}

\subsection{Proof of Theorem \ref{thm:inductive_bias_recover}}
\label{app:thm:inductive_bias_recover}
\begin{proof}
  We follow the proof for Theorem \ref{thm:bias_mf_recover}. Again let $X=F_u D F_v^T$.  
  First, we show that \eqref{eq:Rbound1} is still true for the inductive case. 
  The only difference here is to show that $\min_{X: \|X\|_*\leq t}R_{\alpha, \rho} \geq \eta \min_{X: \|X\|_*\leq t} R_{l_\alpha, \rho}(X) $ because now all the $(i,j)$ elements are dependent. 
  However, as discussed in the previous proof, if we treat each $(i,j)$ independently, 
  the optimal value for each $(i,j)$ elements will be 
  \begin{equation*}
    Z_{ij} = \begin{cases}
      >q \text{ if } A_{ij}=1, \\
      \leq q \text{ if } A_{ij}=0. 
    \end{cases}
  \end{equation*}
  By assumption we know there exists an $D$ with $\|F_u D F_v^T\|_* = \|D\|_*\leq t$ and $X=F_u DF_v^T$ satisfies the above
  condition. Therefore the value of $\min_{X:\|X\|_*\leq t}R_{\alpha, \rho}$ still takes the same value 
  with Theorem \ref{thm:bias_mf_recover}. 
  On the other hand, since now we enforce a more strict constraint that $X=F_u D F_v^T$, 
  Theorem \ref{thm:bias_mf_recover} gives an upper bound of $\min_{D:\|D\|_*\leq t}R_{l_\alpha, \rho}(F_u D F_v^T)$. 
  Therefore, equation \eqref{eq:Rbound1} still holds. 
 
  We then also have
  \begin{equation}
R_{l_\alpha, \rho}(F_u D F_v^T) - \min_{D} R_{l_\alpha, \rho}(F_u D F_v^T) \leq
6\frac{\sqrt{\log(2/\delta)}}{\sqrt{mn}(1-\rho)}
+\frac{4t\sqrt{\log 2d}}{\sqrt{mn}\sqrt{1-\rho}} \X_u \X_v
    \label{eq:Rlbb_1}
  \end{equation}
  using the similar proof to Theorem \ref{thm:inductive_shift_1}. 

  Combining \eqref{eq:Rlbb_1}, \eqref{eq:Rbound1} and Theorem \ref{thm:naga}, the proof is complete.  
\end{proof}


\end{document}